\documentclass{article}

\usepackage{arxiv}

\usepackage[utf8]{inputenc} 
\usepackage[T1]{fontenc}    
\usepackage{hyperref}       
\usepackage{url}            
\usepackage{booktabs}       
\usepackage{amsfonts}       
\usepackage{nicefrac}       
\usepackage{microtype}      
\usepackage{lipsum}
\usepackage{graphicx}
\title{A literature review on current approaches and applications of fuzzy expert systems}

\author{
  Mina Rajabi \\
  Department of Computer Science\\
  Yazd University\\
  \texttt{rajabi.mina@yazd.edu.ir} \\
   \And
 Saeed Hossani \\
  Department of Computer Science\\
  Islamic Azad University of Taft\\
    \texttt{hossani@iau.taft.edu.ir} \\
    \And
 Fatemeh Dehghani \\
  Department of Information Science\\
  Islamic Azad University of Ardakan\\
    \texttt{dehghani@iau.ardakan.edu.ir} \\
}

\begin{document}
\maketitle

\begin{abstract}
The main purposes of this study are to distinguish the trends of research in publication exits for the utilisations of the fuzzy expert and knowledge-based systems that is done based on the classification of studies in the last decade. The present investigation covers 60 articles from related scholastic journals, International conference proceedings and some major literature review papers. Our outcomes reveal an upward trend in the up-to-date publications number, that is evidence of growing notoriety on the various applications of fuzzy expert systems. This raise in the reports is mainly in the medical neuro-fuzzy and fuzzy expert systems. Moreover, another most critical observation is that many modern industrial applications are extended, employing knowledge-based systems by extracting the experts' knowledge.

\end{abstract}


\section{Introduction}
\begin{figure}
\centering
  \includegraphics[width=0.7\linewidth]{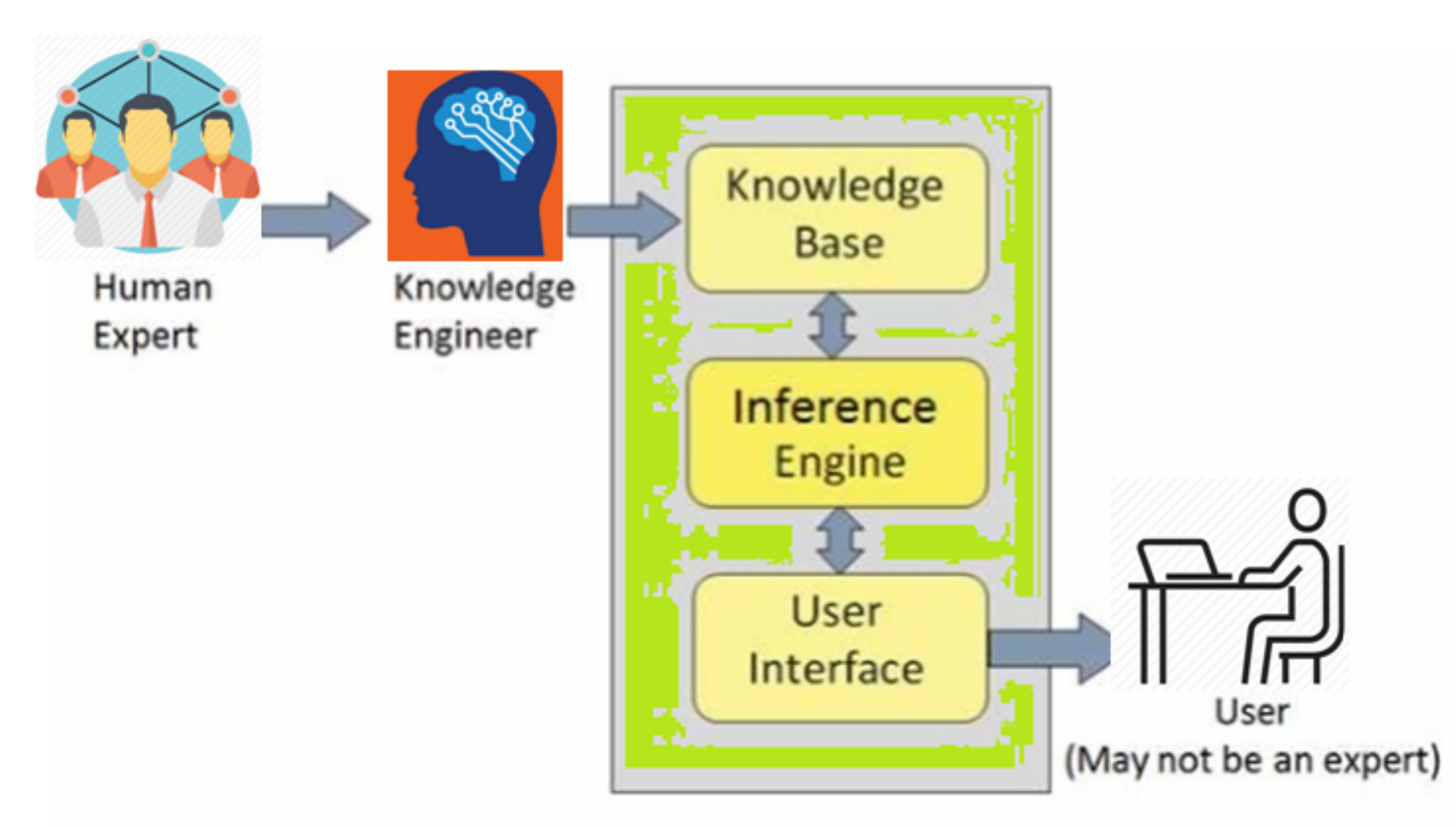}
  \caption{The scheme of an expert system }
  \label{fig:ES}
\end{figure}

Introduction
Expert systems are a division of Artificial Intelligence (AI) which advances widespread use of techno-scientific human expertise to resolve semi or ill-structured problems where there is not a specific guaranteed for solving the algorithm. The expert systems have been characterised as a smart program that applies knowledge and inference steps to handle difficult problems to need important human expertise for their solutions \cite{feigenbaum1981expert}.  The structure of the expert system can be seen in Figure \ref{fig:ES}.

A fuzzy expert system is an expert system, which is composed of fuzzification, inference rules, knowledge database, and defuzzification parts, and employs fuzzy logic rather of the Boolean logic to deliberate about data in the inference mechanism. This procedure is adopted to explain decision-making problems, where there is no accurate algorithm exists, although instead, the problem solution can be estimated heuristically, which is depends on experts in the form of If-Then rules. A fuzzy expert system can be fully satisfied to the problem, which shows uncertainty emanating from fuzziness, uncertainty or subjectivity. 

\begin{figure}
\centering
  \includegraphics[width=\linewidth]{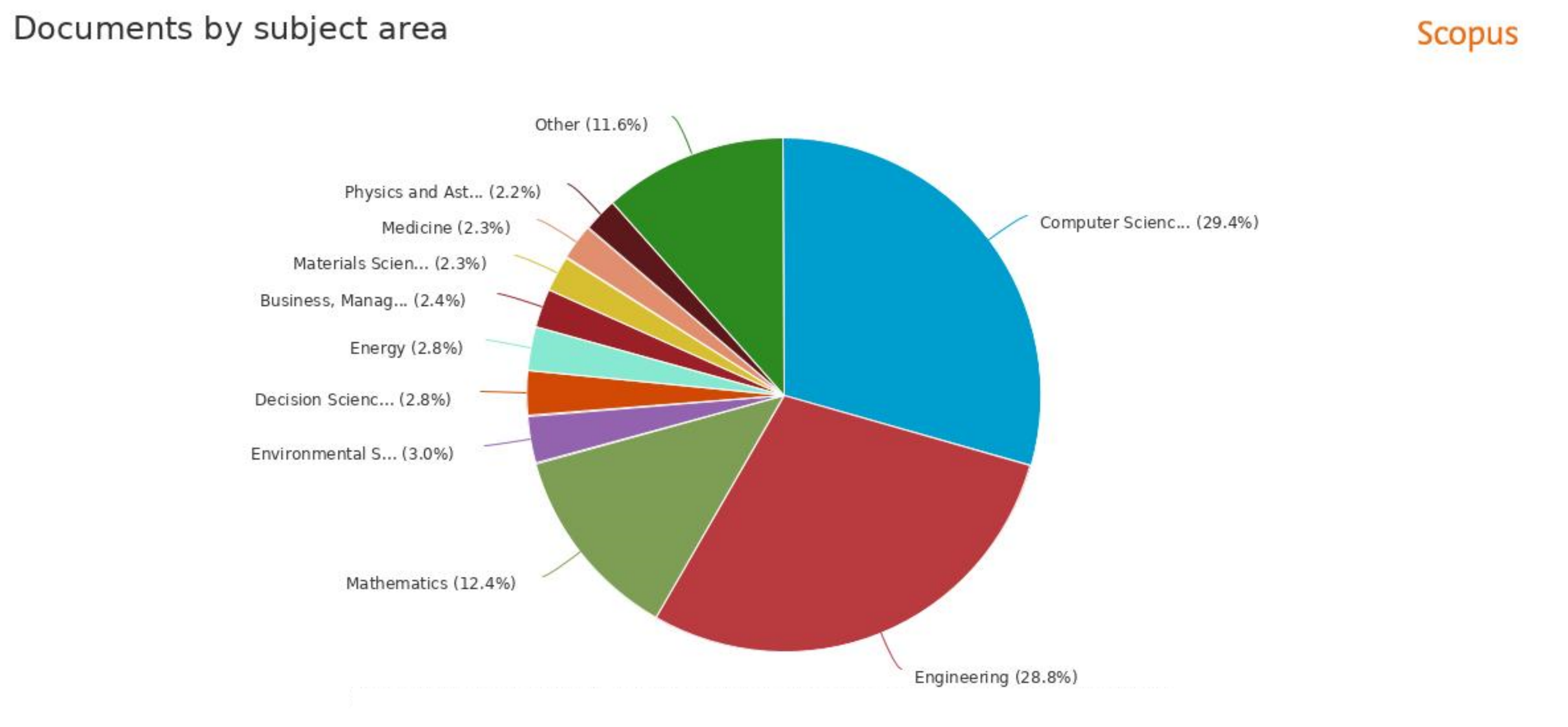}
  \caption{The categorized relevant articles about fuzzy expert systems approaches and applications between in different subjects. The statistical results are from Scopus database \cite{burnham2006scopus} }
  \label{fig:Scopus-Analyze-Subject}
\end{figure}

Despite some fluctuations in the growth trend of fuzzy expert systems publications, the overall trend has been excessively increasing between1983 and 2018 with the number from less than ten papers to more than 1000 publications per year. This statistical information is presented in Figure \ref{fig:Scopus-Analyze-Subject}. 

The main objective of this article is to discuss the principal benefits and shortcomings of the current approaches and theories for improving and modelling fuzzy expert systems in different areas such as industry, medicine, business and so on. With regard to achieving such aims, a comprehensive survey of the relevant publications is reviewed.
The remainder of this paper is prepared as regards. Section 2 describes the fuzzy expert systems in detail. In Section 3, the application of the fuzzy expert systems is investigated by comparing their performances in different aspects of medicine. Furthermore, industrial and customized fuzzy expert systems are surveyed comprehensively in Section 4 and 5, respectively. Finally, conclusions are outlined in Section 6.

\begin{figure}
\centering
  \includegraphics[width=0.6\linewidth]{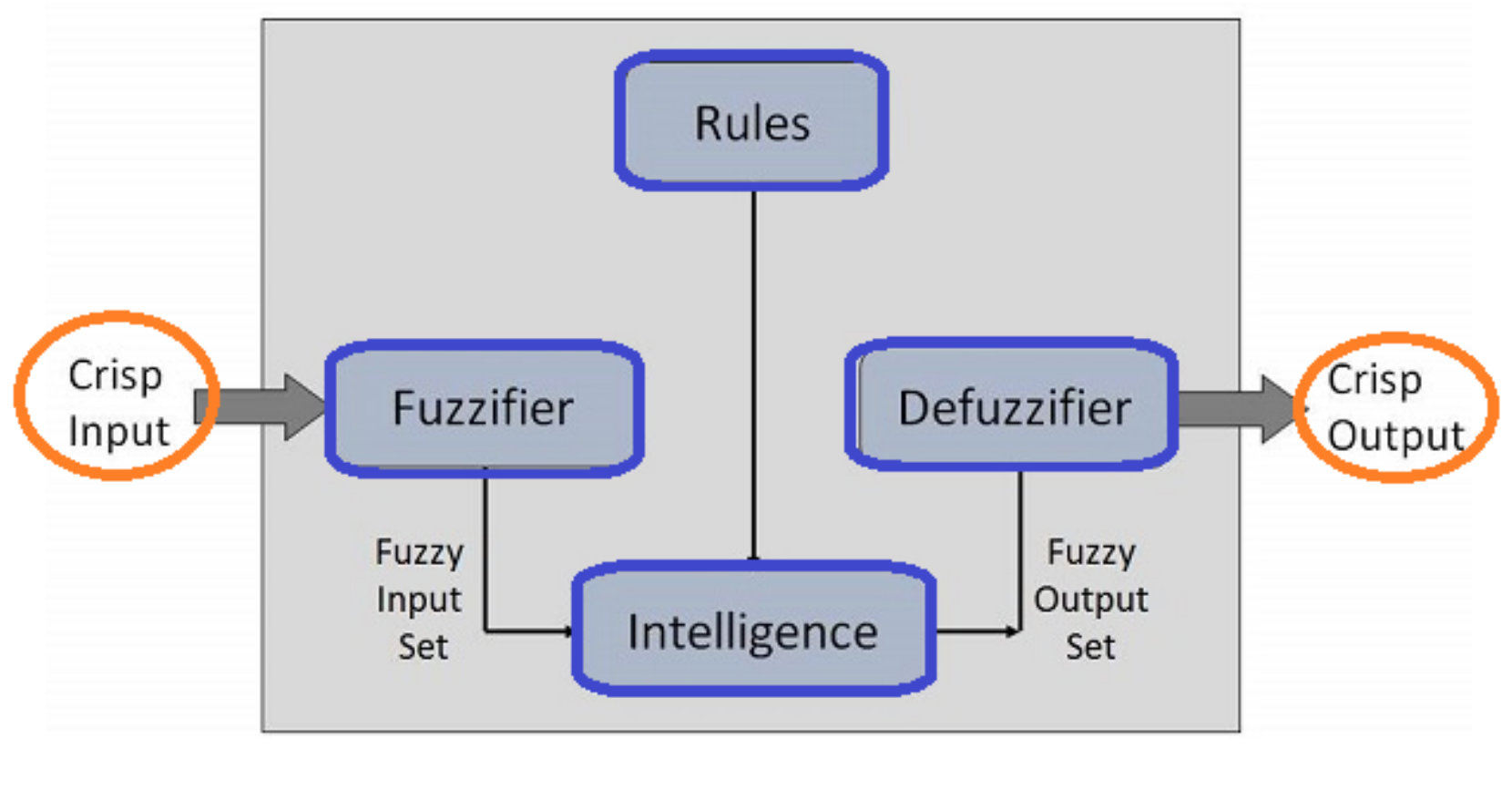}
  \caption{The landscape of a fuzzy expert system }
  \label{fig:FES}
\end{figure}

\begin{figure}[h]
\centering
  \includegraphics[width=0.8\linewidth]{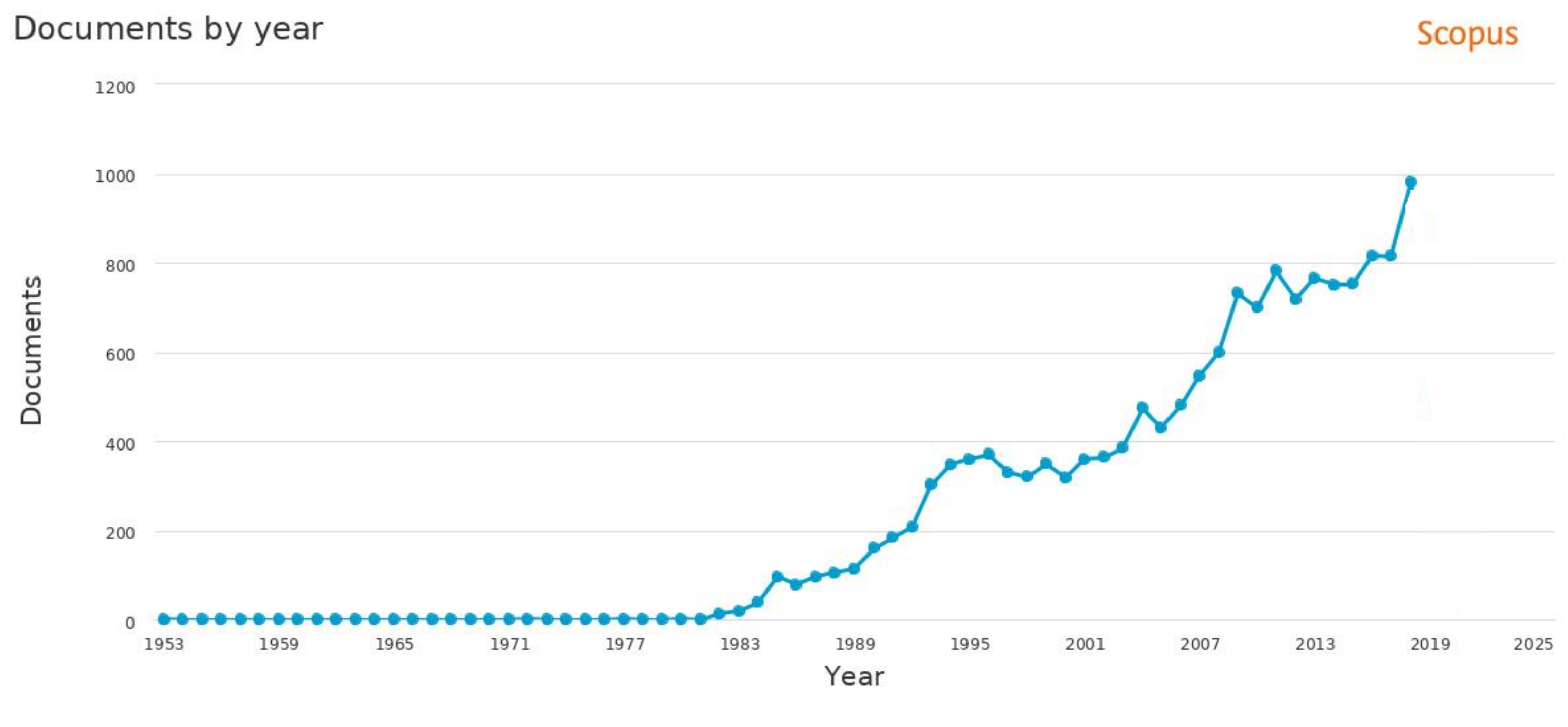}
  \caption{The published number of relevant articles about fuzzy expert systems approaches and applications between 1953 and 2018. The statistical results are from Scopus database \cite{burnham2006scopus} }
  \label{fig:Scopus-Analyze-Year}
\end{figure}

\begin{figure}
\centering
  \includegraphics[width=0.6\linewidth]{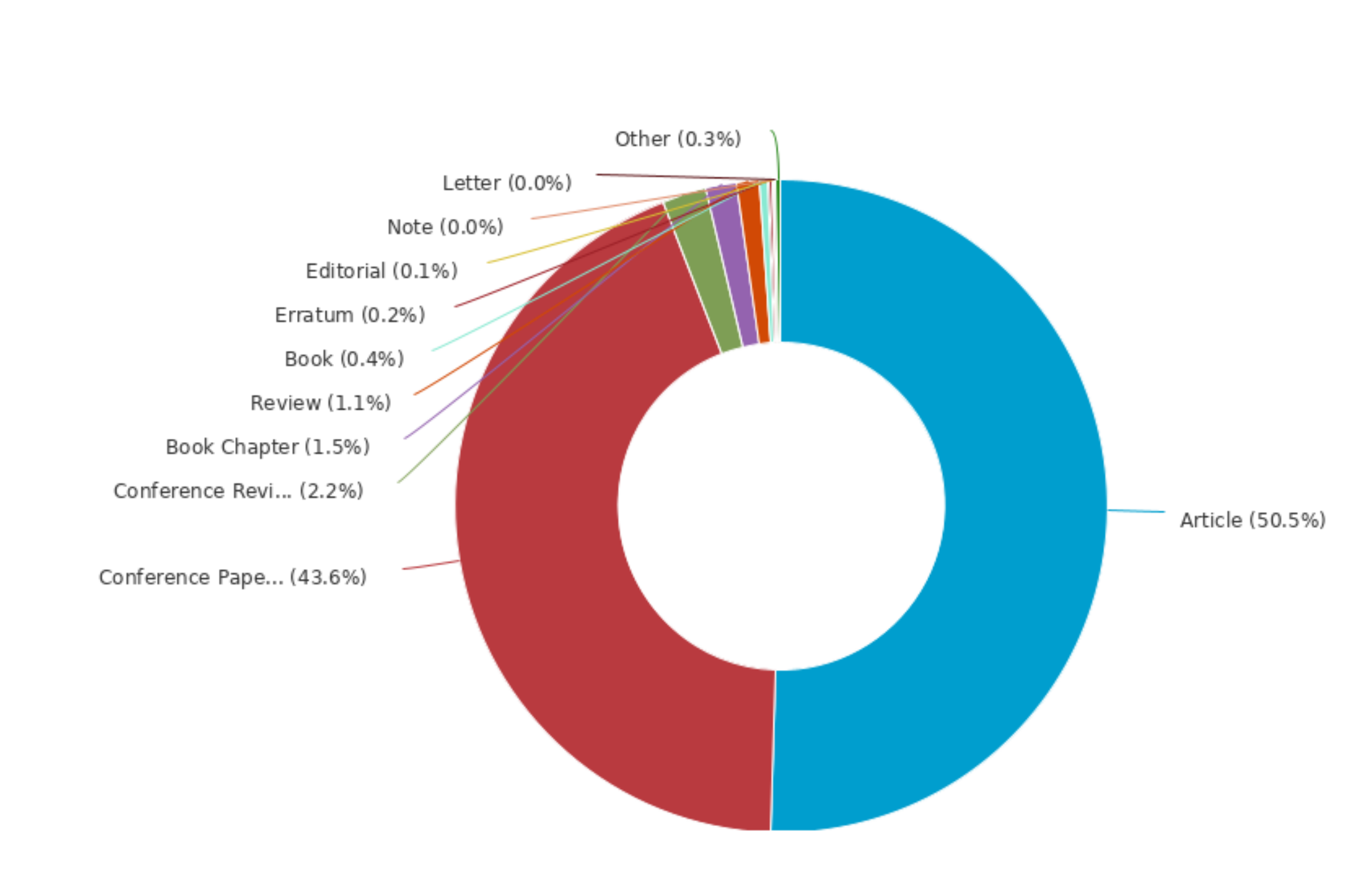}
  \caption{The grouped published articles about fuzzy expert systems approaches and applications based on the types. The statistical results are from Scopus database \cite{burnham2006scopus} }
  \label{fig:Scopus-Analyze-Doctype}
\end{figure}
\section{Fuzzy Expert Systems (FES)}
\label{sec:fes}
Initially, Zadeh \cite{zadeh1996soft} introduced the main theory of fuzzy logic as an approach for interpreting human knowledge that is not precise and well-defined. Figure \ref{fig:FES} shows the fundamental form of a fuzzy logic system.  The process of fuzzification interface converts the crisp information into fuzzy linguistic values by various kinds of membership functions. The fuzzification can be regularly required in a fuzzy expert system considering the input values from surviving detectors are always deterministic numerical values. The inference generator demands fuzzy input and rules, and then it will produce fuzzy productions. Considerably, the fuzzy rule base should be in the figure of “IF-THEN” rules, including linguistic variables. The last part of a fuzzy expert system can be defuzzification which has the responsibility of performing crisp yield operations. The landscape of the fuzzy expert system can be represented in Figure \ref{fig:FES}.

According to the latest statistical SCOPUS information about publishing the research results of fuzzy expert systems in different domains, the maximum percentages are taken into account Computer Sciences, Engineering and Mathematics at $29.4\%,28.8\%$ and $12.4\%$. 

\begin{figure}
\centering
  \includegraphics[width=0.5\linewidth]{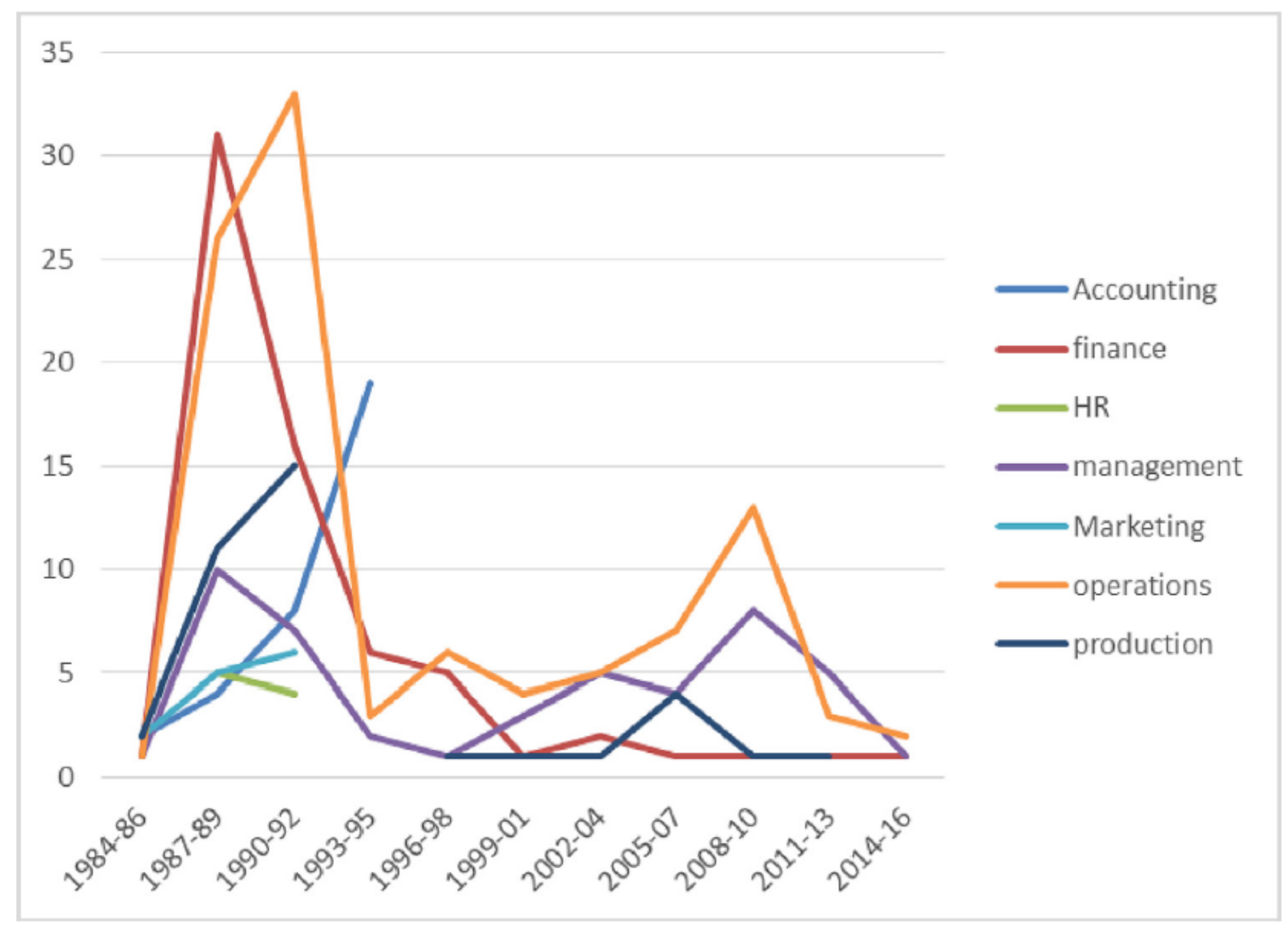}
  \caption{The categorized publications about fuzzy expert systems approaches and applications based on the areas from \cite{wagner2017trends} }
  \label{fig:Analyze}
\end{figure}

Figure \ref{fig:Scopus-Analyze-Subject} shows a detailed percentages of different publications areas in Fuzzy Expert system field. In addition, around half of publications in this scope is the  journal research papers and $43.6\%$ of all publications is related to conference papers (Figure \ref{fig:Scopus-Analyze-Doctype}). However, The importance of fuzzy system applications has been falling down in some specific areas in the last two decades substantially like Accounting, finance, management, marketing, operations and production because of the obtained new research achievements and representing new tools for dealing with uncertainty and fuzziness. These results can be shown in Figure \ref{fig:Analyze}.    

\begin{figure}
\centering
  \includegraphics[width=0.3\linewidth]{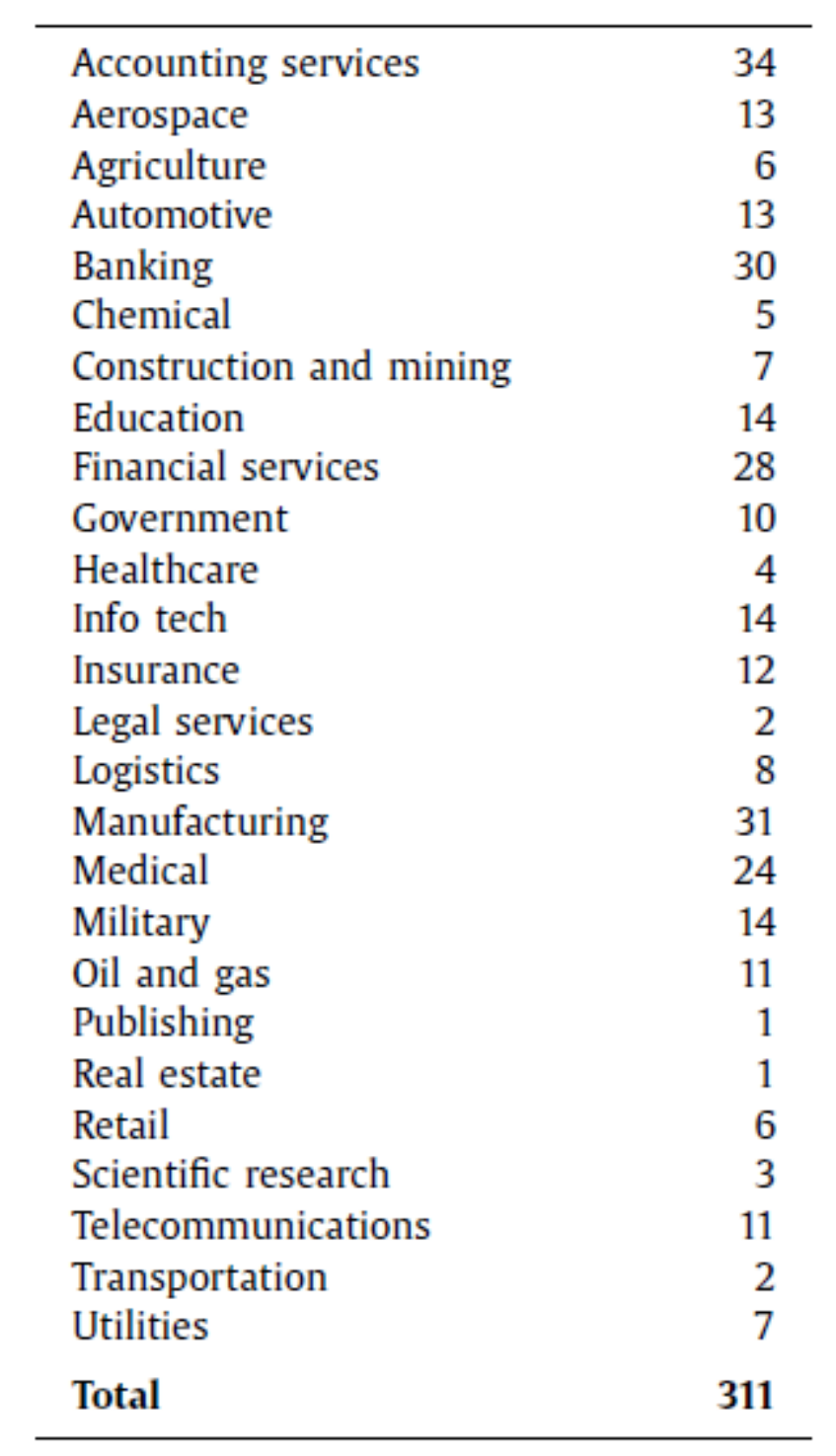}
  \caption{The number of publications about industrial fuzzy expert systems applications from \cite{wagner2017trends} }
  \label{table:num}
\end{figure}

Table \ref{table:num} also shows the number of relevant publication in the same period of the time in different research fields from \cite{wagner2017trends}. The highest ranks are devoted to the Accounting services, Banking and Manufacturing publications.     

\section{ Medical fuzzy expert systems applications}
\label{sec:medical}
In the last three decades, Fuzzy rule-based systems is a subsidiary of Artificial intelligence fitted of interpreting complicated medical data. Their potential to employ significant relationship within a data set has been used in the diagnosis, treatment and predicting consequence in various clinical outlines. A survey of different artificial intelligence methods is exhibited in this part, along with the study of critical clinical applications of expert systems. The ability of artificial intelligence systems and has been explored in almost every field of medicine. Artificial neural network and knowledge based systems were the most regularly accepted analytical tool while additional AI systems such as evolutionary algorithms, swarm intelligence and hybrid systems have been handled in various clinical environments. It can be concluded that AI and expert systems have a high potential to be employed in almost all fields of medicine. Table \ref{table:medicine} shows the application of practical AI techniques such as fuzzy sets, neural networks, evolutionary algorithms, swarm intelligence for diagnosing a wide set of diseases. 
\begin{table}
\begin{center}
    \begin{tabular}{ p{3cm} | p{7cm} | p{3cm} | p{1cm} }
    \hline \hline
    \textbf{Authors} & \textbf{Methods} & \textbf{Disease} & \textbf{Year} \\ \hline
     Polat et al. \cite{polat2006hepatitis}&artificial immune recognition system with fuzzy resource allocation  & Hepatitis disease  & 2006 \\ \hline
   Polat et al. \cite{polat2006diagnosis} &artificial immune recognition system and fuzzy weighted pre-processing &heart disease & 2006 \\ \hline
    Polat et al. \cite{polat2006new} & Artificial immune recognition system (AIRS) with fuzzy weighted pre-processing & ECG arrhythmia &2006  \\ \hline
    Polat et al. \cite{polat2007expert} & adaptive neuro-fuzzy inference system  & diabetes disease &2007  \\ \hline
      Sahan et al. \cite{csahan2007new}& fuzzy-artificial immune system  & Breast Cancer & 2007 \\ \hline
    
Polat et al. \cite{polat2008artificial}     & Artificial immune recognition system with fuzzy resource allocation mechanism classifier & EEG signals & 2008 \\ \hline
     Neshat et al. \cite{neshat2008designing,neshat2008fuzzy,neshat2010hopfield,neshat2014diagnosing,Neshat2013asurvey}& Bayesian parametric method and Parzen window non parametric method, Fuzzy Expert System, Hopfield Neural Network and Fuzzy Hopfield Neural Network & liver disorders  &2008, 2009, 2010, 2013, 2014  \\ \hline
     Neshat et al. \cite{neshat2009designing, neshat2012hepatitis,neshat2009feshdd}& Adaptive Neural Network Fuzzy System, Hybrid Case Based Reasoning
and PSO, Fuzzy expert system & Hepatitis B & 2009, 2012 \\ \hline
     Adeli et al. \cite{adeli2010fuzzy}& Fuzzy Expert System & Heart Disease &2010 \\ \hline
     Baig et al. \cite{baig2011anaesthesia}&Fuzzy logic  & Anaesthesia &2011  \\ \hline
     Shabghahi \cite{shabgahi2011cancer}& Genetic Fuzzy System & Cancer Tumor & 2011 \\ \hline
    Yun et al. \cite{zou2012research} & computer-assisted intelligent system & sexual precocity & 2012 \\ \hline
     Zolnoori et al. \cite{zolnoori2012fuzzy}& Fuzzy Rule-Based Expert System & Asthma & 2012  \\ \hline
     Sachidanand et al. \cite{singh2012diagnosis}& Fuzzy Inference System & Arthritis & 2012 \\ \hline
     Karimpour et al. \cite{karimpour2014fuzzy}& Fuzzy Modeling & Drug Prescription &2014 \\ \hline
     Gayathri et al. \cite{gayathri2015mamdani}& Mamdani Fuzzy Inference system & Breast cancer & 2015  \\ \hline
     Economou et al. \cite{economou2015exploiting}&  Expert Systems &Cardiology  &2015  \\ \hline
     Neshat et al. \cite{neshat2015new}& Fuzzy expert system &Skin Disease &2015 \\ \hline
     Fialhoa et al. \cite{fialho2016mortality}& probabilistic fuzzy systems & septic shock patients &2016 \\ \hline
     Saikia et al. \cite{saikia2016early} & Fuzzy Inference System & Dengue Disease & 2016 \\ \hline
     Alqudah \cite{alqudah2017fuzzy}& Fuzzy expert system & Heart disease &2017  \\ \hline
      Hassan et al. \cite{}& Fuzzy Soft Expert System & Coronary Artery & 2017 \\ \hline
     Sharma et al. \cite{sharma2018fast}& hierarchical fuzzy system & Autism & 2018 \\ \hline
     Karegowda et al. \cite{karegowda2017knowledge}& Knowledge Based Fuzzy Inference System & Diffuse Goiter &2018  \\ \hline
     Sajadi et al. \cite{sajadi2018diagnosis} & fuzzy rule-based expert system & hypothyroidism & 2018 \\ \hline
     Moya et al. \cite{moya2019fuzzy}& Fuzzy‐description logic & elderly rehabilitation  & 2019 \\ \hline
     Guzman et al. \cite{guzman2019optimal}& Genetic Design of Type-1 and Interval Type-2
Fuzzy Systems & Blood Pressure &2019  \\ \hline
      Nguyen et al. \cite{nguyen2018design}& Medical Expert System & Tuberculosis &2019  \\ \hline
     Hadzic et al. \cite{hadvzic2019expert}& Expert System & Anesthesia &2020  \\ \hline

    \end{tabular}
    \caption{A survey of the medical applications of expert rule-based systems and AI techniques.}
\label{table:medicine}
\end{center}
\end{table}
\section{The applications of industrial fuzzy expert systems}
\label{sec:industrial}
We can see many applications of fuzzy expert system in developing the performance of productions in various industries. As the number of new commodities improved by novel technologies had grown, the significance of the new technology products commercialization had become critical to manufactures in the thriving delivery of worthwhile new merchandises and services. The progress factors for commercialization of new products were classified and analyzed which factors have to be principally contemplated. Concerning the Delphi method \cite{cho2013development, keliji2018investigating}, recognized four determination areas and additional prioritized the sixteen parameters following a hierarchy model designed by fuzzy AHP (analytic hierarchy process) method. The FAHP was accompanied by 111 R\&D and enterprise authorities working at the world’s major players in the machinery industry; using the preferences of success agents originated by FAHP. On the other hand,  Traditional mathematical modelling procedures have been utilized to calculate the numbers that have the smallest imprecision. However, not only Type-1 fuzzy sets have applied widely in industrial problems, but also type-2 fuzzy sets have been manipulated for observations where the imprecision rate is comparatively high. A comprehensive survey application of type-1 and type-2 fuzzy systems was published by Dereli et al. in \cite{dereli2011industrial}.  In this respect, the observations were using both types of fuzzy systems in the industry can produce robust and reliable solutions for the problems.  

Analysing and managing the risk of the industrial environments for employees is another attractive research subject of fuzzy expert system. For instance the dropping of height can be one of the most important safety concerns in the building industry, because of the high number of fatal injuries. Scaffolds are a foremost case and have one of the most crucial injury scales. Consequently, introducing the deterrent measures and strategies can be crucial. A hybrid approach of an Adaptive Neural Network-based Fuzzy Inference System (ANFIS) and a safety inspection checklist was introduced by \cite{jahangiri2019neuro} to distinguish hazard factors and prognosticate the risk of slumping from a scaffold on construction sites. That method was able to identify and evaluate critical circumstances that have the most significant impact on fall peril. The hybrid system was perceived to beat the regression method in estimating the fall risk.

Generating a fuzzy system \cite{wang2017knowledge} and hybrid fuzzy decision support system \cite{nourian2019fuzzy} to decrease the risk in gas industries connected to gas transmission services was proposed. The advanced fuzzy expert system was programmed by C, CLIPS and is joined with MATLAB for calling fuzzy membership functions. The expert system gives more than one thousand rules based on the experts' idea to limit the pressure fall and the quality damage of gas or shutting off gas flow which consequently develops gas flow permanence. The recommended hybrid system could reduce the risk of hazardous situations, such as leakage and corrosion. 

In a recent reference \cite{islam2019knowledge}, limiting cost overruns of such infrastructure schemes of power plants can be a project management issue. For dealing with the issue, a knowledge-based expert system was introduced using a fuzzy canonical model (FCM) which combines the fuzzy group decision-making approach (FGDMA). This approach developed the fuzzy-Bayesian models applications for cost overrun risks evaluation in a complicated and uncertain environment. The practical results showed the considerable performance of hybrid method compared with traditional inference methods. Besides, Mayadevi et al. \cite{mayadevi2014review} had reviewed the published articles on expert system applications in power plants deeply. 

In civil engineering, the applications of fuzzy expert systems are noteworthy. The concrete mix design \cite{neshat2011designing, neshat2012predication, Mehdi_Neshat_2012} is so tricky and results in imprecision. Applying fuzzy sets is a technique to describe a variety of uncertainty which is acceptable for experts. Therefore, a fuzzy expert system was implemented to determine the concrete mix designs in a representative form quickly. The system inputs include Slump, Maximum Size of Aggregate (Dmax), Concrete Compressive Strength (CCS) and Fineness Modulus (FM). The empirical outcomes revealed that the MSE of estimated compressive strength for FIS is $6.43\%$, the minimum error of which is $4.73\%$.

Construction industry is a major field of that have absorbed many applications of fuzzy system publications \cite{rathore2015framework}. In this content, Experts have focused on understanding and recognising a proper organization structure that meets the needs of the construction industry. The need for a functional organization structure that is more flexible and can tackle the challenges of the competitive industry led to the adoption of
Matrix Management. Regarded as one of the ideal structures
for effective control of decentralized systems, it enables organizations
to be more adaptive to trends. Nevertheless, Matrix
organizations have also been incorporated with properties of
uncertainty, confusion and inefficiency due to struggles between
different functional roles. This paper aimed to understand and identify the reasons of conflicts between different functional parts within an organization. The introduced model strived to recognise the roots of conflict in the first steps. It was fundamental research that intended to improve a tool that helps the best management in distinguishing potential struggle states and supports in making critical decisions

In the meantime, The fuzzy controllers widely applied in the control
industry are of the PID model. The performance of these
controllers based on the values of the Proportional, Integral
and Derivative, (P, I and D) parameters. Tuning of these
parameters is a tiresome task which demands experience and
expert knowledge of the calibration engineer. In \cite{sam2017performance}
suggested a fine-tuning procedure for these parameters applying
fuzzy logic. The PID Controller was originally tuned using Ziegler
Nichols Closed Loop method and Relay auto-tuning and then
finely tuned using fuzzy logic. The result was drastically developed when fuzzy logic was utilised for fine-tuning, and the results proving the equal are simulated applying MATLAB/Simulink.

A fuzzy control scheme tackled the heating control problem. The recommended Fuzzy logic systems were used to keep the room temperature \cite{kobersi2013control} in the needed range by determining the maximum and minimum temperature. One of the principal Artificial Intelligence means for automatic control was the application of fuzzy logic controllers, which were fuzzy rule-based systems, including expert knowledge in the form of linguistic rules. The rules were usually created by an expert who connected the facts with the outcomes. Fuzzy Logic was a model for an alternative design methodology used in developing both linear and non-linear systems for embedded control. By applying fuzzy logic, designers recognised lower development costs, better features and end-product performance. In control systems, there were several generic systems and techniques which were encountered in all fields of industry and technology. From the dozens of methods to manage any system, it turned out that fuzzy was often the best way because of being cheaper and faster. One of the successful application that applied fuzzy control was a heating control system. The reference explained the idea of using simulation as a mechanism for performance validation and energy  analysis of heating systems.

Automatic vehicles have attracted considerable attention in research and industry \cite{dadgar2016simulating} because of  their potential advantages in driving without passengers. Utilizing computer simulation in motion and speed control of the vehicles can decrease faults and costs. Fuzzy logic implements a diverse approach for issues that need to be tested. The approach was focused on what the system should do, not on how the works were done. Applying fuzzy logic was simple and easily capable to solve complex issues, which were not solvable with conventional mathematical techniques, in less time. This logic worked as an expert’s knowledge. In \cite{dadgar2016simulating} for controlling a machine without passengers, it was aimed to apply steering wheel control and pressing the accelerator pedal to get to the destination of a fuzzy controller so that the machine made decisions and determined the correct path while getting close to the obstacles.
A good application of fuzzy system in rail industry can be \cite{friedrich2019fuzzy,gul2018fuzzy}. Each disruption can have a significant impact on system operation because of infrastructure constraints in the railway transportation system. It is possible that dispatching actions can affect the disrupted traffic. Traffic reconfiguration may cause a quick system recovery, although it may also affect cascade failures and safety infringement. The study of a recovery procedure is challenging to do because of parameters that should be considered. The structure of timetable is a crucial determinant that affects the transportation system robustness. A significant number of possible cases of train schedules in the timetable makes it impossible to tackle this factor. Hence, the paper aimed to present a way which supported the dispatching decision-making method. The article began with a literature review involved the decision-making process and the system recovery. Then, essential indicators of correct train traffic were classified, such as several delayed trains, average train delay, average stop loss and the possibility of more disruptions. Due to the challenge to connect the indicators, the fuzzy logic was selected to create an evaluation approach. The membership functions and the decision rules were placed together with experts from the railway industry. The research was implemented to determine the usability of the approach. As a result, a discussion about more research about the analyzed topic was implemented.

The study in \cite{hemayatkar2019developing} presented the scenarios applying the unknown factors in the business environment, and it also chooses the most reliable approaches of the organization for dealing with the formulated scenarios utilising the fuzzy information revealed by the experts in the application of fuzzy inference system in carpet industry. The aimed in \cite{hemayatkar2019developing}  presented a technique allowing the scenario programmers to apply robustness philosophy applying the scenario planning potentials and fuzzy inference system at the decision-making step of the general process of strategy formulation. The approach supported the strategic managers of the organizations to manage their business future and allowed them to choose their robust scenario in the uncertain market. After the introduction of the robust strategic planning methodology and explaining its different levels, the chosen approaches of them compared by performing the strategic planning approach in a practical case. The outcomes of the research  measured as a case study for the carpet industry.

Food industry is another branch of most important applications of fuzzy sets for modelling ambiguous relationships among systematic variables. Halalan Toyyiban \cite{zakaria2019fuzzy} issues confirmed the necessity of additional protection for good quality assurance food and should be used in the food industries. All the food ingredients must be safe, hygiene and nutritious to be applied by the users. Many chemical ingredients have added in the food production process in improving the food characteristics and function as a stabilizer, artificial colour, preservatives, artificial sweetener. This article showed Halal Food Additive utilizing Fuzzy Expert Systems (FES) to discover the Halalan Toyyiban safety rating for food additives according to consumers’ past encountered record in allergy complication. The importance of this plan is to implement the users with health awareness to stop the health risk that made by these food additives. The study concentrated on 42 kinds of Halal food additives and 12 kind symptoms of allergies that were usually faced by the users.

\section{  Customised fuzzy expert systems}
\label{sec:others}
Except for the mentioned topics, fuzzy systems which are worked based on the extracted knowledge of human expertise have been used in a wide range of applications that called customised in this paper. 

Kawamura and Miyamoto \cite{kawamura2003condition} proposed a new method for improving a concrete bridge rating expert system for deteriorated concrete bridges, built from multi-layer neural networks. The system assesses the performance of concrete bridges based on a simple visual inspection and technical specifications. The principal aim of applying the neural network is that it implements fuzzy inference in the network, promotes refinement of the knowledge base by use of the back-propagation technique, and limits not only the expert system inference tool but also the knowledge base after machine learning from shifting a black box. 

In another prior work of combined FMEA and fuzzy system for evaluating the risk of warehouse operations discussed in \cite{ustundag2012risk}. A recent application of fuzzy system in the modelling of the risk management, Ghoushchi et al. \cite{ghoushchi2019extended} showed the most pivotal  elements of risk management in any organization are identification and prioritization of failure modes in a system and plan for improving procedures. While one of the most commonly used techniques for prioritization of the failures is conventional Failure Mode and Effects Analysis (FMEA). Although the extensive applications of this technique in several industries, FMEA is correlated with some deficiencies that can make unrealistic results. In this article, a proposed method is performed in three stages to overcome some of the shortcomings of the FMEA method. In the first stage, FMEA is applied to recognize the failure modes and assign values to the Risk Priority Number (RPN) determinant factors. In the second stage, the Fuzzy Best-Worst Method (FBWM) according to the experts’ ideas, is utilized to estimate the weights of these elements. In the third stage, the results of the previous stages are applied as a basis to prioritize the failures using the proposed Multi-Objective Optimization by Ratio Analysis based on the Z-number theory (Z-MOORA). With consideration of assigning different weights to the RPN determinant factors and counting possibilities of them, the Z-number theory is applied in this method to reach reliability in different failure modes. The aimed method was performed in the automotive spare parts industry, and the results show a full prioritization of the failures compare to  other conventional approaches such as FMEA and fuzzy MOORA. 

Fuzzy expert structures can be a good tool for modelling the qualifications of the marketing productions. In the developed world \cite{neshat2011fhesmm} , the most significant factor to have a prosperous trade and production is boosting the satisfaction of customers. For increasing the profit of a company, new marketing approaches and supervising the marketing choices play a pivotal role. The paper \cite{neshat2011fhesmm} examined an expert system by four foremost principles of marketing (price, Product, Place and Promotion) and their composition with a fuzzy system and helping from the experiences of marketing specialists. In comparison with  the other applied systems, It had specific properties such as investigating and
extracting different areas in which influence the customers’ satisfaction directly or indirectly as input parameters, applying the information of experts to create inference system rule, composing the results of five fuzzy expert systems and determining ultimate result (customer’s satisfactions) and eventually making a high
function expert system on the management and conducting the managers to do successful marketing in active markets. 
In other recent publication in 2016 \cite{neshat2016designing}, Neshat et al. proposed a fuzzy framework for achieving the success in business and attendance in the real world markets, it is crucial to outperform the competitors to occupy a more significant market share. To gain customer satisfaction from products can be the first step of achievement in business. Studying the different factors included in growing the level of customer's satisfaction and studying in this field had made improvements in various businesses. In \cite{neshat2016designing}, an adaptive neuro-fuzzy inference system (ANFIS) and a fuzzy inference system (FIS) were implemented for marketing mix model by utilizing the P4 principle (price, product, place, promotion) and by joining it with the marketing experts' knowledge, significant results were gained applying ANFIS. The system as an advisor with high precision can decrease human errors and had an essential role in decision making by corporate managers. The consequences of the two systems were compared, and it represented that ANFIS had a better performance than FIS with mean accuracy of $98.6\%$ and $87.25\%$, respectively.

Evaluating the quality of the proposed services in the organisations play a significant role to develop their performances. One recent study which was done in the United States of America (USA) \cite{engin2014rule} showed that there are more than 15 million college students. Educational advising for studies and scholarships is commonly accomplished by human advisors, causing an enormous managerial workload to faculty members and other university staffs. That study illustrated and explained the improvement of two educational expert systems at a private international university. The first expert system was a course advising system which suggests courses to undergraduate students. The second system advised scholarships to undergraduate students according to the eligibility of them. While there have been informed systems for course advising, the literature did not include any references to expert systems for scholarship advice and eligibility checking. Consequently, the scholarship advisor that they improved was first of its type. Oracle Policy Automation (OPA) software was applied too for implementing and testing of both systems. 
With regard to evaluation of the service quality, Pourahmad et al. \cite{pourahmad2016using, pourahmad2012service} showed that the application of a fuzzy system and hybrid fuzzy expert system to assess and evaluate the level of quality services of four various university libraries in the North Khorasan province of Iran. The statistical groups included various students from several branches, and they were selected as samples. For accumulating data, the survey method was used; meanwhile, data collection tool and particular questionnaire were applied since each of the four elements for quality estimation of services was calculated using the LibQUAL tool. The mean total services for libraries of the university of North Khorasan were negative in terms of service fitness gap, which means that libraries were not able to perform the minimum expectation of their users. For all library services, the gap was negative. In other terms, libraries are far from compensating the expectations of students associated with the most acceptable (maximum) level of services.

One of the most interesting applications of fuzzy systems can be modelling and simulating the dynamic climate models. The climate change can be the most extensive scale problem \cite{koca2019causes, neshat2013rainfall} and the well-known troubling question in the current world. Human-caused global warming is driving climate change impacting both human and natural systems on all continents and across the oceans . Human-caused global warming results from the increased use of fossil fuels in transportation, manufacturing and communications. The higher emission of greenhouse gasses has increased the global temperature on Earth’s surfaces. Koca et al. \cite{koca2019causes} explained a Fuzzy Inference System (FIS) to predict the connection between the causes and effects of climate change. The proposed Fuzzy Inference System presented a baseline for recognizing the causes of climate change, with the help of expert knowledge, which indicated the correlation between multiple environmental factors. According to a Multiple-Input-Multiple-Output Mamdani Fuzzy Inference System, the consequences of climate change can be measured by obtaining a predicted result with considering a variety of climate scenarios. Here, CO2 (Carbon Dioxide), Global Temperature Changes, Snow Cover, Percentage of Forest lands, and Net Radiation are observed as inputs; While the estimated effects are Changes of Ozone Layer, Arctic sea ice decline and ice loss. Three group functions of the “generalized bell function” type have been estimated for each input and output. Moreover, Koca et al. displayed that multiple inputs and multiple outputs in a Fuzzy Inference System can influence the way for understanding climate models. 

For modelling the environmentally disasters such as forest fires \cite{neshat2016hybrid}, heavy storms \cite{oladokun2017measuring}, earth quacks \cite{andalib2016fuzzy}, FIS can play an important role. Indeed, Fire is one of the essential factors destroying forest ecosystems \cite{neshat2016hybrid} which can lead to adverse economic and social outcomes. Fast detection can be a useful determinant in controlling this devastating phenomenon. In \cite{neshat2016hybrid} that was proposed at designing a hybrid fuzzy expert system in order to predict the size of forest fires efficiently and accurately. The data were obtained from the real dataset named forest fire in University of California (UCI). The introduced system is a hybrid of six fuzzy inference systems with acceptable performances based on their results. The precision of predicting the size of the fire was $81.2\%$.
Furthermore, Expert systems had a notable role in better doing of complex tasks and giving advice to the experts because expertism was specialised knowledge \cite{neshat2016recognising}. The expert systems were used to solve the problems for which there was not an exact knowledge and a particular algorithm. Understanding the atmospheric phenomena and their role in human life were the most significant and affecting issues in human societies. In meteorology, it was essential to recognise the kind of clouds. By observing from the Earth's surface (seeing the bottom view of the cloud) and utilising satellites (seeing the top view of the cloud), we can classify the variety of clouds. A fuzzy inference system with the specialists' knowledge of meteorology was planned in the paper, and its objects were the detection of the cloud type by extracting knowledge from satellite images of the cloud upper portions. The applied data were extracted from the reliable website of UCI called cloud dataset. The data set was gathered by Philip Collard in two ranges of IR and VISIBLE.  This system concluded the type of cloud with a precision rate of $88.25\%$ $\pm 0.5$ and based on experts' idea; the results were proper and satisfactory.

Meanwhile, fuzzy system can be combined with other approaches to create an effective tool for improving the practical results.In this way, the study \cite{adeli2012comparison} proposed and investigated developing a Linux-base operating system with the LFS approach. Firstly, we have a quick glimpse of the history of Linux and the reasons for its significance in the current world. Next, we explain different kinds of Linux Distributions and their internal structures and describe how to improve a Linux Distribution in different ways. LFS technique is compared with another technique called "Remaster", and its benefits and drawbacks are explained. This technique was examined gradually  from the start to reaching a Linux-based operating system. 

Expert systems are used in agriculture \cite{dubey2013literature}, which assists the farmers to make the appropriate decisions for higher crop production. Expert systems for pest management and crop protection compose a critical class of agricultural expert systems. Knowledge of entomology, plant pathology, nematology, weeds and nutritional disorders and a diverse number of methods used, are involved in integrated pest management and crop protection. Uncertainty is faced during the time of sowing,  diagnosis of insect, weed management, disease and nutritional disorders, storage, marketing of the product. This uncertainty is combined with the fact that many agricultural decision- making actions are often inexplicit or based on intuition. Fuzzy logic is applied to manage imprecision, vagueness and insufficient knowledge. Fuzzy logic allows expert systems to perform optimally with uncertain or ambiguous data and knowledge. Fuzzy expert systems apply fuzzy logic rather than classical Boolean logic. They are orientated towards numerical processing. A review  \cite{dubey2013literature} of several fuzzy expert systems in agriculture during the last two decades was done.

\section{  Conclusions}
\label{sec:con}
This article reviews diverse recent published papers around the cloverleaf of practical fuzzy neural networks, expert systems specialities and particularly focuses on modern leanings in the applications and developments of fuzzy expert systems. Additionally, 
some relevant review papers are evaluated concerning the fuzzy inference system frameworks and strategies, categorized applications and modern tools for implementing such platforms. We summarize our results in some tables and three charts. According to the findings, an uptick trend in the new number of publications can be seen, which can be an implication of augmenting reputation on the fuzzy expert systems. Our outcomes reveal a downward trend in the number of publications which are individually on neural fuzzy expert systems. Many various applications have developed for fuzzy expert systems, but the principal improvement is related to the industrial applications. 
This increment can be related to the appearance of modern applications which need reforms in the approaches and the structures of the frameworks. Finally, we note that the majority of proper publications which have the highest impact factor have been published in specific conferences and journals like Expert Systems and Application (H-Index=162) and Knowledge-Based Systems (H-Index=94).

\bibliographystyle{unsrt}  
\bibliography{references}

\end{document}